\documentclass[runningheads]{llncs}

\usepackage{graphicx}
\usepackage{amsmath,amssymb}
\usepackage{multirow}
\usepackage{url}

\begin{document}

\title{DDR-Net: Haze-Aware Dual-Domain Refinement for Single-Image Dehazing}
\titlerunning{Haze-Aware Dual-Domain Refinement Network for Single-Image Dehazing}

\author{
  Xinye Zheng\inst{}\and
  Ye Yu\thanks{Corresponding author}\and
  Qiang Lu\inst{}\and \\
  Jinsheng Luo\inst{}\and
  Yiran Cui\inst{}\and
  Yongbin Cheng\inst{}
}

\authorrunning{X. Zheng et al.}
\institute{School of Computer Science and Information Engineering, \\  Hefei University of Technology, \\ Hefei, China\\
\email{yuye@hfut.edu.cn}}

\maketitle

\begin{abstract}
Single-image dehazing aims to recover clear scenes from haze-degraded images. It remains challenging due to the atmospheric scattering and the complexity of real-world haze distributions. Although recent end-to-end networks have achieved promising performance, two issues still limit their effectiveness: insufficient feature refinement at the bottleneck stage and weak local structural representation in encoder-decoder architectures. Thus, we propose a Haze-Aware Dual-Domain Refinement Network (DDR-Net) for single-image dehazing. Our method is built upon three modules: Haze Prior Extractor (HPE) provides multi-scale haze-aware priors by operating directly on downsampled hazy images; Detail-Enhanced Blocks (DE Blocks) serve as the core feature extraction units, capturing multi-scale structural information and enhancing edge and texture recovery via gradient-aware convolutions; and Spatial-Frequency Bottleneck Refinement (SFBR) at the bottleneck jointly exploits spatial and frequency information to refine bottleneck features. DDR-Net achieves more effective feature representation and reconstruction for haze removal. Extensive experiments on real-world benchmarks demonstrate that our method outperforms existing dehazing approaches. It achieves competitive performance on synthetic datasets.
\end{abstract}


\section{Introduction}

Single-image dehazing aims to recover a clear scene from a haze-degraded image. It serves as a critical preprocessing step for a variety of practical applications, including autonomous driving and remote sensing. In hazy environments, particles such as aerosols and water droplets scatter and absorb incident light, which leads to low contrast and color distortion. Meanwhile, these degradations introduce a negative impact on visual quality and downstream vision tasks such as object detection, semantic segmentation, and scene understanding. Thus, developing effective image dehazing methods is important for both academia and industry.

\begin{figure}[!t]
    \centering
    \includegraphics[width=\textwidth]{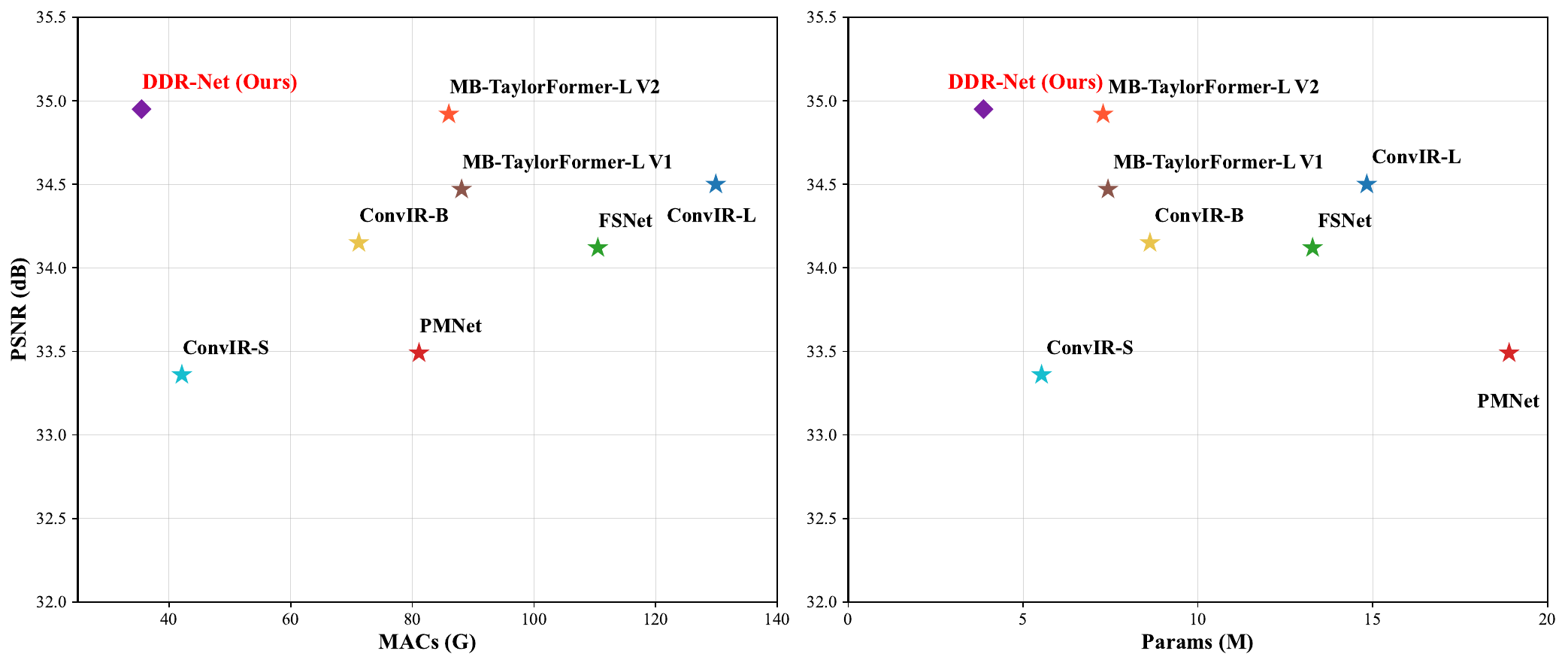}
    \caption{Comparison of representative SOTA dehazing methods on the Haze4K dataset. Left: MACs (G) vs. PSNR (dB); Right: Params (M) vs. PSNR (dB). Our DDR-Net achieves competitive dehazing performance with lower computational cost and fewer parameters than recent SOTA methods.}
    \label{fig:haze4k_compare}
\end{figure}

Early dehazing approaches relied on handcrafted priors, such as Dark Channel Prior (DCP)~\cite{he2010single} and Color Attenuation Prior (CAP)~\cite{zhu2015fast}. Although these methods can achieve satisfactory results under certain assumptions, their performance often degrades in complex real-world scenarios due to limited generalization ability and sensitivity to parameter settings. With the rapid development of deep learning, CNN-based dehazing methods have significantly advanced the field~\cite{cai2016dehazenet,ren2016single,qin2020ffa,li2017aod,guo2022image,chen2024dstvis}. DehazeNet~\cite{cai2016dehazenet} and MSCNN~\cite{ren2016single} employed convolutional networks to estimate transmission maps, while AOD-Net~\cite{li2017aod} reformulated the atmospheric scattering model for end-to-end image restoration. Furthermore, FFA-Net~\cite{qin2020ffa} introduced feature attention for more effective representation learning. More recently, Transformer-based methods such as Dehamer~\cite{guo2022image} and Rest\-ormer~\cite{zamir2022restormer} have further improved dehazing performance by modeling long-range dependencies.

Recent studies~\cite{wang2023frequency,xu2022multi,chen2025mcll,niu2025sad,yi2026mrtp,yu2024night} show that low-frequency components characterize global haze while high-frequency details are critical for restoration. Several methods explore joint spatial-frequency learning. SAD-Net~\cite{niu2025sad} uses wavelet transform and difference convolutions to sharpen edges. JSFC-Net~\cite{zhang2025beyond} employs Fourier transform for global receptive fields and mutual modulation. However, these methods rely on handcrafted masks and lack balancing of spatial and frequency information.

To address the aforementioned issues, we propose a Haze-Aware Dual-Domain Refinement Network (DDR-Net) for single-image dehazing. We design three key modules to enhance feature representation and information flow. We introduce a lightweight Haze Prior Extractor (HPE) that directly provides early‑stage haze distribution based on downsampled hazy images. To strengthen local structural representations, we develop Detail-Enhanced (DE) Blocks as the core feature extraction units in our encoder‑decoder. At the bottleneck stage where features possess the largest receptive field and the richest semantics, we propose the Spatial-Frequency Bottleneck Refinement (SFBR) Module to perform collaborative feature refinement in spatial and frequency domains. As shown in Fig.~\ref{fig:haze4k_compare}, DDR-Net achieves competitive performance on the Haze4K dataset with lower computational cost.

Our key contributions are summarized as follows:
\begin{itemize}
\item We propose a lightweight HPE that injects multi‑scale haze‑aware priors into the encoder.
\item We design DE Blocks as the feature extraction unit, which capture multi‑scale structural information and enhance edge and texture recovery.
\item We introduce SFBR for spatial‑frequency bottleneck  refinement, suppressing spatial redundancy and enhancing high-frequency details with low overhead.
\item We build DDR-Net by integrating HPE, DE Blocks, and SFBR into an encoder‑decoder architecture, achieving competitive dehazing performance on synthetic and real‑world benchmarks.
\end{itemize}

\section{Related Work}

\subsection{CNN-Based Methods}
Early CNN-based dehazing methods followed two directions: estimating intermediate variables in the atmospheric scattering model or directly learning a mapping from hazy images to clean images. DehazeNet~\cite{cai2016dehazenet} estimated the transmission map with a lightweight CNN and recovered the clean image based on the atmospheric scattering model. AOD-Net~\cite{li2017aod} further reformulated the physical model into an end-to-end framework. Multi-scale designs, such as MSCNN~\cite{ren2016single} and GridDehazeNet~\cite{liu2019griddehazenet}, aggregated features with different receptive fields.

Recent CNN-based dehazing networks have explored improved convolutional operators and feature modulation mechanisms~\cite{chen2024dea,cui2024revitalizing,liu2024sdcnet,shen2026efficient,Yu2023RIC-NVNet}. DEA-Net~\cite{chen2024dea} introduced enhanced convolution and content-guided attention for local representation. IRNeXt~\cite{cui2024revitalizing} combined multi-scale design with local attention to emphasize high-frequency information. SDCNet~\cite{liu2024sdcnet} adopted deformable convolution and a dual-branch architecture to reduce the negative impact of hazes. FSPN~\cite{wen2025single} introduced a fine‑scale perception network employing multi‑branch convolution and maximum pooling for detail extraction, while SEMN~\cite{shen2026efficient} proposed a synergic expert modulation mechanism that integrates Mixture-of-Experts (MoE)~\cite{jacobs1991adaptive} to enhance context modeling. These methods demonstrate the effectiveness of CNNs in detail restoration, while long-range dependency modeling under complex haze conditions remains challenging.

\subsection{Transformer-Based Methods}
Early Transformer-based methods are limited by the quadratic computational complexity of self-attention with respect to spatial resolution. To improve efficiency, SwinIR~\cite{liang2021swinir} introduced a hierarchical Swin Transformer~\cite{liu2021swin} with shifted-window attention. Uformer~\cite{wang2022uformer} further developed a U-shaped architecture based on locally enhanced window Transformer blocks, and incorporated depth-wise convolutions into the feed-forward network to strengthen local context modeling. Restormer~\cite{zamir2022restormer} redesigned self-attention by computing cross-covariance across channels rather than spatial positions.

Recent Transformer-based dehazing methods have focused on improving both haze-aware modeling and computational efficiency~\cite{song2023vision,liu2021swin,guo2022image,qiu2023mb,jin2025mb,lin2026derestormer}. DehazeFormer~\cite{song2023vision} improves the standard Swin Transformer~\cite{liu2021swin} by redesigning the normalization layer, the activation function, and the spatial information aggregation for better image dehazing. Dehamer~\cite{guo2022image} introduced a transmission-aware 3D position embedding and feature modulation strategy, which enables adaptive feature refinement under different haze densities. To reduce computational overhead while maintaining global receptive fields, MB-TaylorFormer~\cite{qiu2023mb} and MB-TaylorFormer V2~\cite{jin2025mb} approximated softmax attention using first-order Taylor expansion and combined it with multi-scale patch embedding based on deformable convolutions. DeRestormer~\cite{lin2026derestormer} incorporates deformable attention and multi‑scale aggregation to reduce computational cost, but its improvement remains incremental. These studies demonstrate the effectiveness of Transformer-based designs for image dehazing. However, balancing global dependency modeling and computational efficiency is still challenging.

\begin{figure*}[!t]
    \centering
    \includegraphics[width=\textwidth]{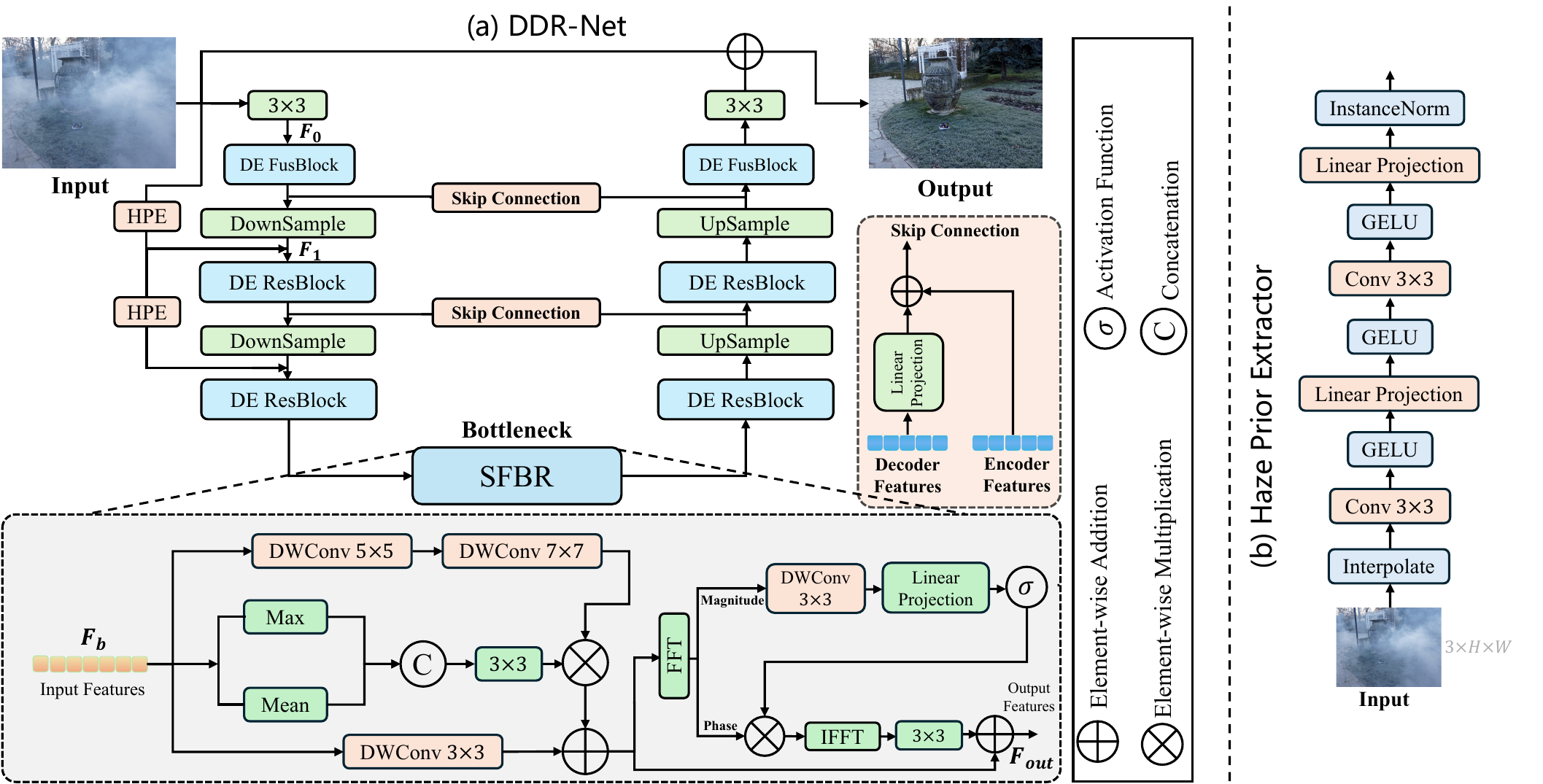}
    \caption{Overall architecture of DDR-Net. (a) The encoder-decoder framework integrates SFBR at the bottleneck. (b) HPE provides complementary haze-aware representations for restoration.}
    \label{fig:architecture}
\end{figure*}

\section{Methodology}

The framework of DDR-Net is illustrated in Fig.~\ref{fig:architecture}. Built upon a U-Net backbone, DDR-Net consists of three proposed modules: HPE for multi-scale haze prior extraction, DE Blocks for haze-aware detail extraction, and SFBR for spatial-frequency feature refinement.

\subsection{Haze Prior Extractor}

To provide the backbone with explicit haze-aware guidance, we introduce a lightweight HPE. Rather than deriving priors from intermediate backbone features, HPE operates directly on the hazy image. As shown in Fig.~\ref{fig:architecture}, given an input hazy image $I \in \mathbb{R}^{3 \times H \times W}$, HPE is applied twice in cascade, with each application performing one downsampling operation. Features from another downsampling branch are then concatenated, forming a multi-scale feature pyramid that explicitly encodes haze distributions at different spatial scales before being injected into the encoder.

HPE uses a shallow convolutional network to extract haze-aware prior features. It uses two \(3\times3\) convolutions to capture local haze patterns and correlated degradations, and two \(1\times1\) convolutions as linear projection to refine intermediate features. GELU activations are inserted between consecutive layers. Let \(Z_s\) denotes the output feature before normalization at scale \(s \in \{1/2,1/4\}\). The final haze-aware prior feature is obtained by instance normalization:
\begin{equation}
H_s(k,u,v)=\gamma_k \frac{Z_s(k,u,v)-\mu_{s,k}}{\sqrt{\sigma_{s,k}^2+\epsilon}}+\beta_k,
\end{equation}
where \(k\) denotes the feature-channel index, \((u,v)\) denotes the coordinate, and \(\mu_{s,k}\) and \(\sigma_{s,k}^2\) are the mean and variance computed over all spatial positions of the \(k\)-th channel in \(Z_s\), respectively.

This combination of convolution, linear projection, and GELU activation preserves the hazy image’s structural properties while capturing haze-related degradations. The subsequent instance normalization further stabilizes feature distributions, enhancing the prior’s robustness to varying haze conditions and improving generalization across diverse scenes.

\begin{figure*}[!t]
    \centering
    \includegraphics[width=\textwidth]{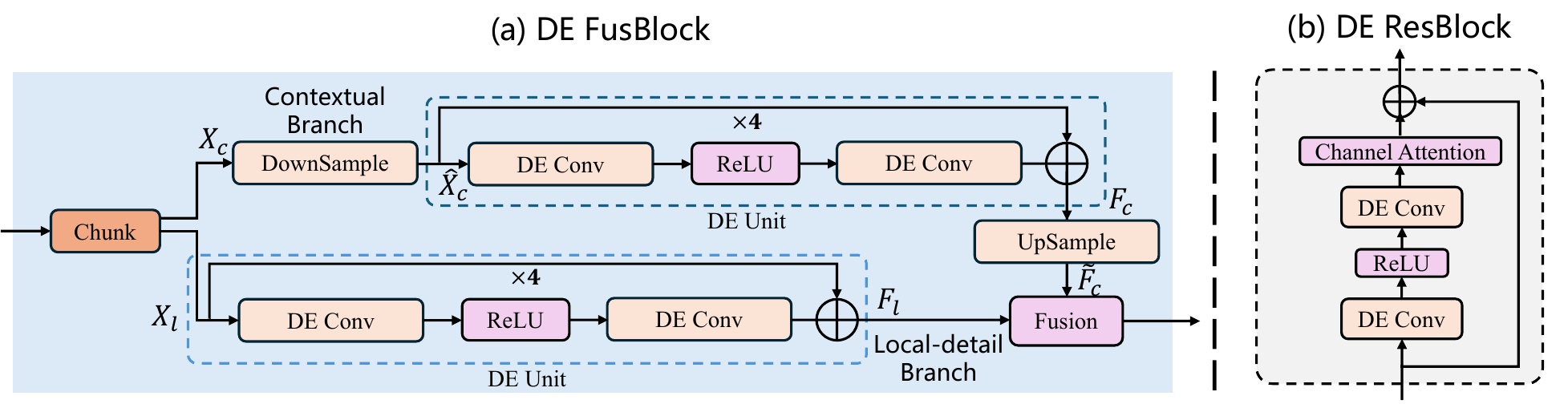}
    \caption{Architectures of the Detail-Enhanced Blocks. (a) DE FusBlock splits the input features into two branches to perform refinement at multiple spatial resolutions. (b) Detailed architecture of DE ResBlock.}
    \label{fig:grresblock}
\end{figure*}

\subsection{Detail-Enhanced Blocks}

We introduce DE Blocks, which consist of two variants: DE FusBlock and DE ResBlock. DE FusBlock is a dual-branch fusion module designed to jointly capture multi-scale contextual information and fine-grained local details for robust haze removal. DE ResBlock enhances haze-relevant features and stabilizes training for effective dehazing. Both blocks leverage DE Conv~\cite{chen2024dea} as their core operator. Specifically, DE Conv comprises five parallel convolution branches: central-difference (CD), horizontal-difference (HD), vertical-difference (VD), angular-difference (AD), and standard convolution (STD), where each branch performs a standard spatial convolution. The detailed designs of DE FusBlock and DE ResBlock are presented in the following subsections.

\textbf{DE FusBlock.} The DE FusBlock is a dual-branch feature decomposition and fusion module tailored for single-image dehazing, as shown in Fig.~\ref{fig:grresblock}(a). Given the shallow feature map \(F_0 \in \mathbb{R}^{32 \times H \times W}\) extracted by the initial convolution, it first employs a channel-wise Chunk operation to split \(F_0\) into two disjoint streams, denoted as \(X_l\) and \(X_c\), for complementary local-detail and contextual representations, respectively.

The contextual branch first projects its split feature \(X_c\) into a compact spatial domain via downsampling, yielding the downsampled feature \(\hat{X}_c\). Reducing the spatial resolution allows subsequent convolutional refinement to operate over a larger receptive field, effectively capturing global haze distributions and large-scale scene structures. The local-detail branch retains the full spatial resolution of \(X_l\) to preserve fine textures and high-frequency details. Both branches are then independently refined by DE Units:

\begin{equation}
F_l = \mathcal{R}(X_l), \qquad
F_c = \mathcal{R}(\hat{X}_c),
\end{equation}
where \(F_l\) and \(F_c\) represent the refined features of the local-detail and contextual branches, respectively. \(\mathcal{R}(\cdot)\) denotes the DE Unit operation, which is composed of two DE Conv layers, a ReLU activation, and a residual connection.

After feature refinement, the low-resolution contextual representation \(F_c\) is restored to its original spatial resolution through an upsampling operation, resulting in the upsampled feature \(\tilde{F}_c\), and aligned with the local-detail branch for cross-resolution feature interaction. The reconstructed contextual feature \(\tilde{F}_c\) and the local-detail feature \(F_l\) are then concatenated along the channel dimension and fused via a learnable convolution layer to produce the final integrated feature \(Y\):

\begin{equation}
Y = W_f * [F_l,\tilde{F}_c] + b_f,
\end{equation}
where \([\cdot,\cdot]\) denotes channel-wise concatenation, and \(W_f\) and \(b_f\) are the weight kernel and bias of the fusion convolution layer, respectively. 

This design explicitly balances high-frequency texture preservation and low-frequency haze distribution modeling, yielding robust multi-scale features that support effective dehazing.

\textbf{DE ResBlock.} The DE ResBlock is a residual building block tailored for haze-relevant feature extraction. As shown in Fig.~\ref{fig:grresblock}(b), it takes an input feature map and processes it through a sequence of operations within a residual skip connection.

First, the input is passed through two stacked DE Conv layers, with a ReLU activation inserted between them. The multi-directional design of the DE Conv enables the block to capture both structural details and haze-induced degradations from different spatial orientations. The output of the second convolution is then fed into a Channel Attention module, which adaptively recalibrates channel-wise feature responses to emphasize haze-relevant information while suppressing redundant features. Specifically, we perform the global average pooling to obtain a channel-wise descriptor:
\begin{equation}
z_c = \frac{4}{H W} \sum_{i=1}^{H/2} \sum_{j=1}^{W/2} X_{channel}(i,j), \quad c = 1,2,\dots,64.
\end{equation}
The descriptor vector \(z \in \mathbb{R}^{64}\) is then passed through a transformation to generate the attention vector \(s\):
\begin{equation}
    s = \frac{1}{1 + \exp\left(-W_2 \cdot \max(0,\, W_1 z)\right)},
\end{equation}
where \(W_1 \in \mathbb{R}^{\frac{64}{r} \times 64}\) and \(W_2 \in \mathbb{R}^{64 \times \frac{64}{r}}\) denote the linear projections implemented via \(1\times1\) convolutions. \(r = 16\) is the reduction ratio. The attention vector \(s \in \mathbb{R}^{64}\) is applied to the feature map via element-wise multiplication. Finally, this refined feature is added to the original encoder feature \(F_1\) through a residual connection.

\subsection{Spatial-Frequency Bottleneck Refinement}

As shown in Fig.~\ref{fig:architecture}, the input to SFBR is a compact feature representation \(F_b \in \mathbb{R}^{128 \times \frac{H}{4} \times \frac{W}{4}}\). While \(F_b\) contains rich contextual information, it still suffers from spatially redundant responses and weakened high-frequency details under non-uniform haze. To address these limitations, SFBR refines the bottleneck feature via spatial and frequency selection.

\textbf{Spatial Selection.} To enhance the discriminability of bottleneck features, we adopt a multi-branch spatial selection mechanism that adaptively recalibrates feature responses across different receptive fields and spatial positions.
Given the bottleneck feature \(F_b\), we process it through three parallel branches:

1. Large receptive field branch:  \(F_b\) is sequentially processed by two depth-wise convolutions (\(5\times5\) and \(7\times7\)) to capture long-range contextual haze patterns.

2. Spatial descriptor branch: Channel-wise max pooling and mean pooling are applied to aggregate information, then concatenated to form a compact spatial descriptor:
\begin{equation}
D(i,j) =
\left[
\max_{1 \leq c \leq C} F_b(c,i,j),\;\;
\frac{1}{C}\sum_{c=1}^{C} F_b(c,i,j)
\right]
\in \mathbb{R}^{2 \times \frac{H}{4} \times \frac{W}{4}},
\end{equation}
where \(F_b(c,i,j)\) denotes the feature value at channel index \(c\) and spatial position \((i,j)\), and \(C\) is the number of channels. This descriptor is then transformed via a \(3\times3\) convolution to produce a spatial attention mask.

3. Residual branch: A \(3\times3\) depth-wise convolution is applied directly to \(F_b\) to preserve fine-grained local details.

The output of the large receptive field branch is multiplied element-wise with the attention mask from the spatial descriptor branch, then added to the residual branch output to produce the spatially refined feature \(F_s\). This design enables the module to suppress redundant spatial responses while enhancing haze-relevant regions, laying a foundation for subsequent frequency-domain refinement.

\textbf{Frequency Selection.} Building on \(F_s\), we perform frequency-domain enhancement to recover high-frequency details such as edges and textures. Specifically, we transform the feature map into the frequency domain via a 2D Fast Fourier Transform (FFT). Since haze degradation mainly affects the magnitude distribution while the phase preserves structural and geometric information, we design a learnable frequency attention weight to adaptively emphasize frequency responses.
Specifically, the magnitude spectrum \(A\) is processed through a \(3\times3\) depth-wise convolution, followed by a linear projection and a sigmoid activation to produce the frequency attention mask \(W_f\).

\begin{equation}
W_f
=
\sigma
\Bigl(
\operatorname{Conv}_{1\times1}
\bigl(
\operatorname{Conv}^{dw}_{3\times3}(A)
\bigr)
\Bigr),
\end{equation}
where \(\sigma(\cdot)\) denotes the sigmoid activation function. The mask is then applied to the magnitude spectrum via element-wise multiplication, while the original phase is preserved. The refined spectrum is then transformed back to the spatial domain via the inverse FFT (IFFT).

Finally, the output of the IFFT is refined by a \(3\times3\) convolution and combined with the original spatially refined feature via a residual connection to produce the final bottleneck output feature \(F_{\text{out}}\).

\begin{table*}[!t]
\centering
\caption{Quantitative comparisons on real-world  and synthetic dehazing benchmarks. Best results are in \textbf{bold}, second-best are \underline{underlined}. ``-'' indicates unavailable results.}
\label{tab:combined_all}
\small
\resizebox{\textwidth}{!}{%
\begin{tabular}{l|cc|cc|cc|cc|cc|cc}
\hline
\multirow{3}{*}{\textbf{Method}}
& \multicolumn{6}{c|}{\textbf{Real-world Datasets}}
& \multicolumn{6}{c}{\textbf{Synthetic Datasets}} \\
\cline{2-13}
& \multicolumn{2}{c|}{\textbf{NH-Haze}}
& \multicolumn{2}{c|}{\textbf{O-Haze}}
& \multicolumn{2}{c|}{\textbf{Dense-Haze}}
& \multicolumn{2}{c|}{\textbf{SOTS-indoor}}
& \multicolumn{2}{c|}{\textbf{SOTS-outdoor}}
& \multicolumn{2}{c}{\textbf{Haze4K}} \\
\cline{2-13}
& \textbf{PSNR} & \textbf{SSIM} & \textbf{PSNR} & \textbf{SSIM} & \textbf{PSNR} & \textbf{SSIM} & \textbf{PSNR} & \textbf{SSIM} & \textbf{PSNR} & \textbf{SSIM} & \textbf{PSNR} & \textbf{SSIM} \\
\hline
MSBDN~\cite{dong2020multi} & 17.97 & 0.659 & 24.36 & 0.749 & 15.13 & 0.555 & 33.67 & 0.985 & 33.48 & 0.982 & 22.99 & 0.850 \\
Dehamer~\cite{guo2022image} & 20.66 & 0.684 & 25.11 & 0.777 & 16.62 & 0.560 & 36.63 & 0.988 & 35.18 & 0.986 & – & – \\
C2PNet~\cite{zheng2023curricular} & 20.24 & 0.687 & 25.20 & 0.785 & 16.88 & 0.573 & 42.56 & \textbf{0.995} & 36.68 & 0.990 & – & – \\
DEA-Net~\cite{chen2024dea} & 20.60 & 0.687 & 25.21 & 0.779 & 16.85 & 0.569 & 40.20 & 0.993 & 36.03 & 0.989 & 33.19 & \textbf{0.990} \\
DEA-Net-CR~\cite{chen2024dea} & 20.66 & 0.689 & 25.23 & 0.781 & 16.87 & 0.571 & 41.31 & \underline{0.994} & 36.59 & 0.990 & 34.25 & \textbf{0.990} \\
ConvIR-S~\cite{cui2024revitalizing} & 20.65 & 0.692 & 25.25 & 0.784 & \underline{17.45} & 0.608 & 41.53 & \underline{0.994} & 37.95 & 0.990 & 33.36 & \textbf{0.990} \\
ConvIR-B~\cite{cui2024revitalizing} & 20.66 & 0.691 & 25.36 & 0.780 & 16.86 & 0.600 & \underline{42.72} & \textbf{0.995} & \textbf{39.42} & \textbf{0.992} & 34.15 & \textbf{0.990} \\
MB-TaylorFormer-B V2~\cite{jin2025mb} & 20.73 & 0.703 & 25.29 & 0.790 & 16.95 & \underline{0.621} & 41.00 & 0.993 & 37.81 & \underline{0.991} & – & – \\
MB-TaylorFormer-L V2~\cite{jin2025mb} & \underline{20.77} & \underline{0.705} & \underline{25.43} & 0.792 & 16.90 & 0.607 & \textbf{42.84} & \textbf{0.995} & \underline{39.25} & \textbf{0.992} & \underline{34.92} & \textbf{0.990} \\
\hline
\textbf{DDR-Net (Ours)} & \textbf{20.89} & \textbf{0.713} & \textbf{25.52} & \textbf{0.792} & \textbf{17.74} & \textbf{0.648} & 42.03 & 0.989 & \textbf{39.42} & \underline{0.991} & \textbf{34.95} & \textbf{0.990} \\
\hline
\end{tabular}%
}
\end{table*}

\section{Experiment}

\subsection{Experiment settings}

\textbf{Real-world Datasets.} We evaluate our model on three benchmarks. O-Haze~\cite{ancuti2018haze} provides 45 pairs of outdoor haze images. Dense-Haze~\cite{ancuti2019dense} presents a more challenging scenario with 33 pairs of outdoor scenes characterized by dense haze. NH-Haze~\cite{ancuti2020nh} includes 55 pairs of outdoor images capturing non-homogeneous haze distributions.

\textbf{Synthetic Datasets.} For synthetic data, we utilize the Synthetic Objective Testing Set (SOTS) from the RESIDE benchmark~\cite{li2018benchmarking}. We report results on both the SOTS-indoor and SOTS-outdoor subsets to cover different depths and lighting variations. We also incorporate the Haze4K dataset~\cite{liu2021synthetic} to  validate the performance.

\textbf{Evaluation Metrics.} We employ Peak Signal-to-Noise Ratio (PSNR) and Structural Similarity Index (SSIM) for the quantitative evaluation. We compute both metrics directly on the RGB color channels of the uncropped images to maintain a fair and consistent comparison with existing techniques.

\subsection{Implementation Details}

We conducted experiments on a single NVIDIA RTX 4090. The encoder consists of three levels with channel numbers of 32, 64, and 128, respectively. SFBR is deployed at the 128-channel bottleneck stage for synchronous spatial feature selection and frequency enhancement. The total parameter of the model is 3.87M.

The model is optimized using AdamW~\cite{loshchilov2017decoupled} optimizer. Hyperparameters $\beta_1$, $\beta_2$, and $\epsilon$ are set to 0.9, 0.999, and $10^{-8}$ with a weight decay of 0.01. Initial learning rate is set to $1 \times 10^{-4}$ and decreased to a minimum of $1 \times 10^{-6}$ using the cosine annealing strategy~\cite{he2019bag} throughout the training process. The batch size is set to 8. We employ a multi-scale L1 loss as the optimization objective. Specifically, we extract the outputs from the three levels of the decoder, upsample them to the original input resolution via bilinear interpolation, and compute the L1 loss against the haze-free ground truth separately. The final loss is the sum of the losses from these three levels. During the preprocessing stage, the training images are resized to a fixed size of $256 \times 256$.

\subsection{Comparison with SOTA models}

\textbf{For quantitative analysis.} Table~\ref{tab:combined_all} presents the quantitative comparisons with existing SOTA methods on real-world and synthetic benchmarks. It achieves state-of-the-art performance across most real-world and synthetic dehazing benchmarks, with the best PSNR and SSIM scores on NH-Haze, O-Haze, Dense-Haze, and Haze4K. On SOTS-outdoor, it attains the highest PSNR and the second-best SSIM, demonstrating strong robustness to diverse haze distributions. While the results on SOTS-indoor are competitive but not top-ranked, this minor variation is likely due to the dataset’s specific indoor haze characteristics, which differ from the non-uniform, complex haze scenarios our architecture is primarily optimized for. Overall, the consistent superior performance across most benchmarks validates the effectiveness and generalization capability of the proposed method.   

\textbf{For computational overhead.} Table~\ref{tab:haze4k_overhead} reports the parameters and MACs of different methods on Haze4K. Our DDR-Net achieves the lowest computational cost among all methods, with only 3.87M parameters and 35.47G MACs. Compared to the second-best methods, it reduces parameters by 30~\% over ConvIR-S (3.87M vs. 5.53M) and MACs by 59~\% over MB-TaylorFormer-L V2 (35.47G vs. 86.0G). These results demonstrate that DDR-Net achieves an excellent balance between restoration quality and computational efficiency.

\begin{table}[!t]
\caption{Quantitative comparison of computational overhead on Haze4K in terms of parameters and MACs.}
\label{tab:haze4k_overhead}
\centering
\footnotesize
\begin{tabular}{l|cc}
\hline
Method & Params & MACs \\
\hline
MSBDN                & 31.35M & \underline{41.5G} \\
FFA-Net              & 4.46M  & 287.8G \\
ConvIR-S             & \underline{5.53M}  & 42.1G \\
ConvIR-B             & 8.63M  & 71.2G \\
ConvIR-L             & 14.83M & 129.9G \\
MB-TaylorFormer-L V1 & 7.43M  & 88.1G \\
MB-TaylorFormer-L V2 & 7.29M  & 86.0G \\
\hline
\textbf{DDR-Net (Ours)} & \textbf{3.87M} & \textbf{35.47G} \\
\hline
\end{tabular}
\end{table}

\begin{figure}[!t]
    \centering
    \includegraphics[width=\columnwidth]{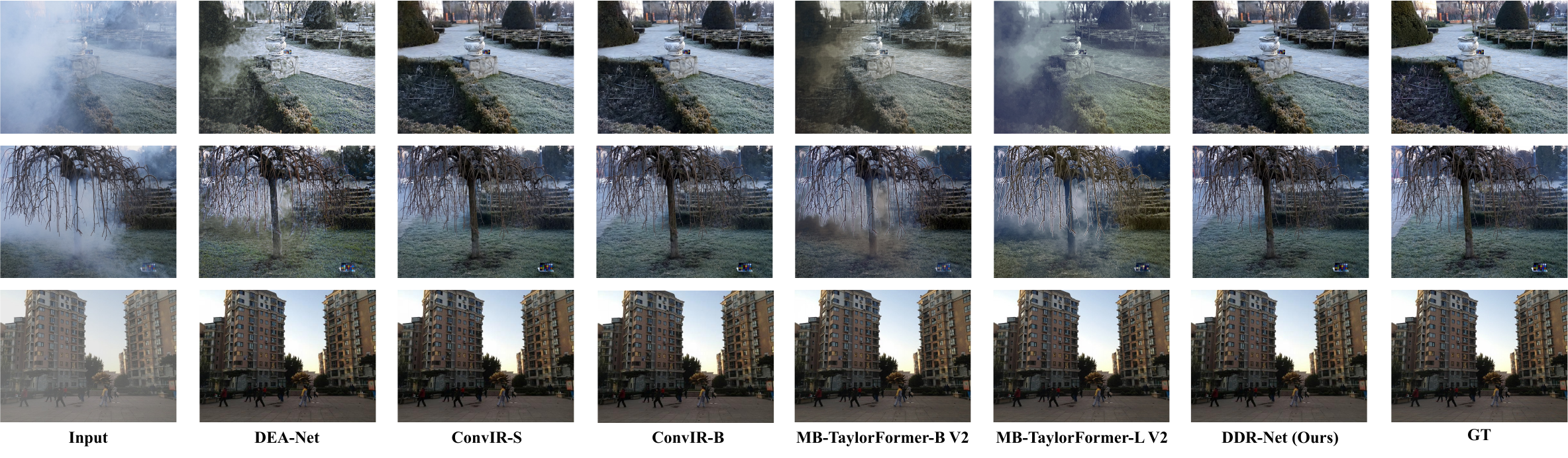}
    \caption{Qualitative comparison on NH-Haze (top two rows) and SOTS-Outdoor (bottom row) datasets. Most models achieve decent results on synthetic data but suffer from color shifts on real-world images. In contrast, our DDR-Net consistently produces images with accurate color and fine details across all scenarios.}
    \label{fig:visual1}
\end{figure}

\begin{figure}[!t]
    \centering
    \includegraphics[width=\columnwidth]{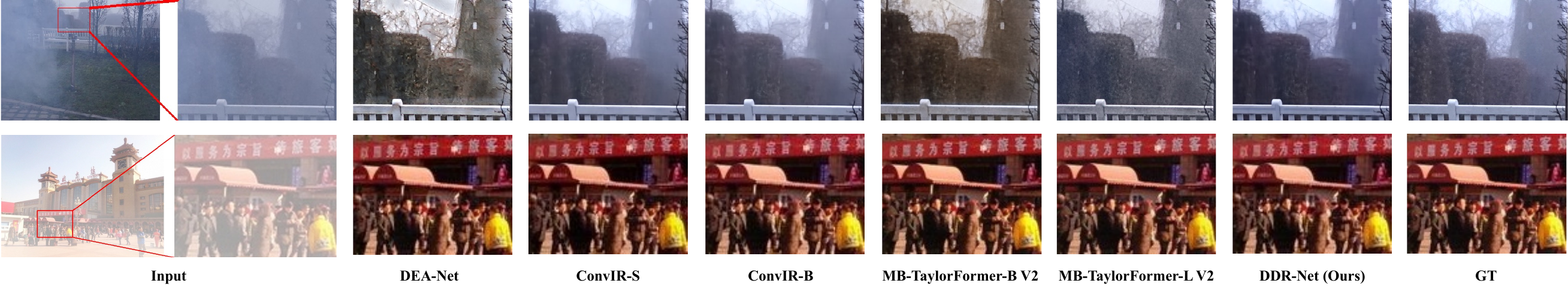}
    \caption{Zoomed-in views of the local regions from NH-Haze (top) and SOTS-Outdoor (bottom). Competing methods produce blurry textures and color distortion on real-world images. DDR-Net consistently recovers fine details and more accurate colors.}
    \label{fig:visual2}
\end{figure}

\textbf{For qualitative analysis.} We present visual comparisons on real-world and synthetic benchmarks to evaluate the dehazing quality of different methods. Figure~\ref{fig:visual1} shows the overall dehazing results on NH-Haze and SOTS-Outdoor. Most competing methods suffer from obvious color shifts and residual haze on real-world scenes. By contrast, our DDR-Net consistently generates clear images with accurate color and fine structural details. Figure~\ref{fig:visual2} provides zoomed-in views of the local regions for a closer inspection. On the real-world NH-Haze example, other methods often yield blurry edges or unnatural colors, while our method preserves sharp textures and correct hues. These qualitative observations align well with the quantitative results reported in Table~\ref{tab:combined_all}, which justifies the effectiveness of our proposed approach.

\subsection{Ablation Study}

\textbf{Evaluation of each component.} To verify the effectiveness of each component in DDR-Net, we conduct ablation studies on NH-Haze. Starting from a baseline encoder-decoder model with standard convolutions, we incrementally integrate the HPE, DE Blocks, and SFBR. As shown in Table~\ref{tab:ablation_components}, the baseline model achieves a PSNR of 18.35 dB and SSIM of 0.660. Adding the HPE provides explicit haze-aware guidance, improving performance to 19.21 dB PSNR and 0.674 SSIM. Replacing standard convolutions with DE Blocks further boosts results to 19.97 dB PSNR and 0.684 SSIM. Finally, deploying the SFBR module at the bottleneck yields the best performance: 20.89 dB PSNR and 0.713 SSIM, as it filters spatial redundancy and enhances high-frequency details. The consistent gains across all components validate their complementary roles in improving dehazing quality.

\textbf{Evaluation of SFBR.} We conduct an ablation study on NH-Haze to verify the spatial and frequency branches in SFBR. As shown in Table~\ref{tab:ablation_SFBR}, the baseline without SFBR achieves 19.97 dB PSNR and 0.684 SSIM. The spatial branch alone improves results to 20.13 dB PSNR and 0.690 SSIM, while the frequency branch alone boosts performance to 20.24 dB PSNR and 0.698 SSIM. The full SFBR design, combining both branches, yields the best performance: 20.89 dB PSNR and 0.713 SSIM. These results confirm that both branches contribute meaningfully, and their integration delivers the strongest improvement, demonstrating their complementary roles in refining bottleneck features.

\begin{table}[!t]
\caption{Ablation study of different components on NH-Haze.}
\label{tab:ablation_components}
\centering
\begin{tabular}{l|cc}
\hline
\textbf{Design} & \textbf{PSNR} & \textbf{SSIM} \\
\hline
Baseline & 18.35 & 0.660 \\
+ HPE & 19.21 & 0.674 \\
+ DE Blocks & 19.97 & 0.684 \\
\hline
\textbf{+ SFBR (DDR-Net)} & \textbf{20.89} & \textbf{0.713} \\
\hline
\end{tabular}
\end{table}

\begin{table}[!t]
\caption{Ablation study of the SFBR module on NH-Haze.}
\label{tab:ablation_SFBR}
\centering
\begin{tabular}{l|cc}
\hline
\textbf{Design} & \textbf{PSNR} & \textbf{SSIM} \\
\hline
w/o SFBR & 19.97 & 0.684 \\
Spatial Branch Only & 20.13 & 0.690 \\
Frequency Branch Only & 20.24 & 0.698 \\
\hline
\textbf{Full SFBR} & \textbf{20.89} & \textbf{0.713} \\
\hline
\end{tabular}
\end{table}

\begin{figure*}
    \centering
    \includegraphics[width=\textwidth]{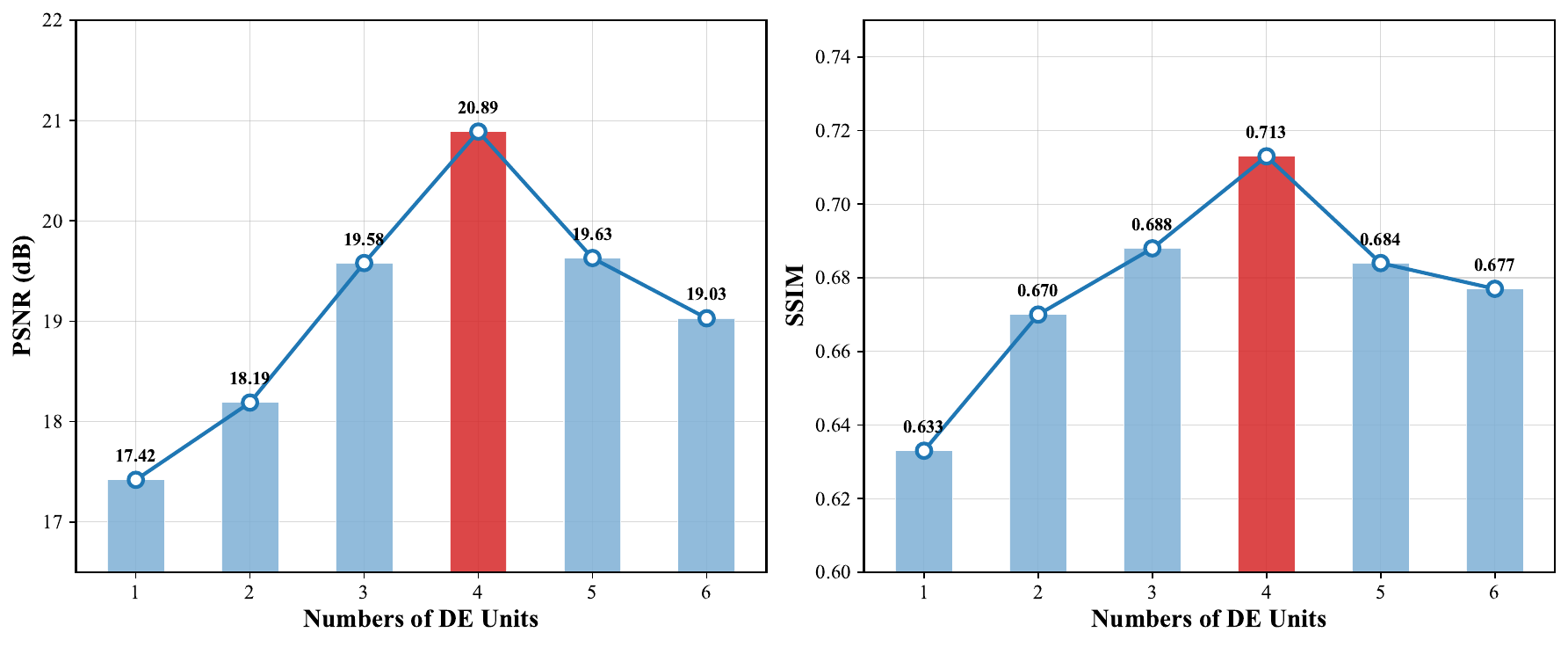}
    \caption{Evaluation on the number of DE Units. The performance exhibits an increasing trend from 1 to 4, reaching its peak values at 4 units. As the number of DE Units increases from 4 to 6, performance gradually decreases.}
    \label{fig:grunits}
\end{figure*}

\textbf{Number of DE Units.} As shown in Fig.~\ref{fig:grunits}, the ablation study on the number of DE Units reveals a clear peaking trend. When the number is set to 1, the model achieves a PSNR of 17.42 dB and an SSIM of 0.633. As this number increases from 1 to 4, both metrics steadily improve, reaching their maximum values of 20.89 dB and 0.713 at 4 DE units. Further increasing the number from 5 to 6 results in a gradual performance decline, dropping to 19.03 dB and 0.677 at 6 units. Therefore, we set the number to 4 as the optimal configuration.

\section{Conclusion}
In this paper, we propose DDR-Net, a haze-aware dual-domain refinement network that addresses the limitations of insufficient bottleneck refinement and weak structural representation in existing dehazing methods. By integrating HPE for multi-scale haze prior extraction, DE Blocks for edge-enhanced feature extraction, and SFBR for spatial-frequency joint refinement, DDR-Net achieves effective haze removal while preserving high-frequency details. Extensive experiments on real-world and synthetic benchmarks demonstrate that DDR-Net consistently outperforms existing methods with lower computational overhead. Ablation studies validate the contribution of each component. Future work will explore video dehazing and unsupervised learning settings to further improve generalization under challenging conditions.


\section*{Acknowledgment}
This work is supported by grants of the National Natural Science Foundation of China (No. 62372153) and the Natural Science Foundation of Anhui Province (No. 2308085MF216).

\bibliographystyle{splncs04}
\bibliography{ref}

\end{document}